\documentclass[letterpaper]{article} 
\usepackage{aaai22}  
\usepackage{times}  
\usepackage{helvet}  
\usepackage{courier}  
\usepackage[hyphens]{url}  
\usepackage{graphicx} 
\usepackage{natbib}  
\usepackage{caption} 
\DeclareCaptionStyle{ruled}{labelfont=normalfont,labelsep=colon,strut=off} 
\usepackage{algorithm}
\usepackage{algorithmic}
\usepackage{amssymb}
\usepackage{subfig}
\usepackage{amsmath}
 
\usepackage{verbatim}
\usepackage{color}

\newcommand{\gray}{\textcolor[RGB]{92,92,92}}
\usepackage{newfloat}
\usepackage{listings}
\floatstyle{ruled}
\newfloat{listing}{tb}{lst}{}
\floatname{listing}{Listing}

\title{Personalized Federated Learning for Multi-task Fault Diagnosis of Rotating Machinery}
\author{
    Sheng Guo\equalcontrib,
    Zengxiang Li\equalcontrib,
    Hui Liu,
    Shubao Zhao,
    Cheng Hao Jin
}
\affiliations{
    \textsuperscript{\rm 1}Digital Research Institute, ENN Group, China\\

    guoshengd, lizengxiang, liuhuiau,  zhaoshubao, jinchenghao@enn.cn
}

\usepackage{bibentry}

\begin{document}

\maketitle

\begin{abstract}
Intelligent fault diagnosis is essential to safe operation of machinery. However, due to scarce fault samples and data heterogeneity in field machinery, deep learning based diagnosis methods are prone to over-fitting with poor generalization ability. To solve the problem, this paper proposes a personalized federated learning framework, enabling multi-task fault diagnosis method across multiple factories in a privacy-preserving manner. Firstly, rotating machines from different factories with similar vibration feature data are categorized into  machine groups using a federated clustering method. Then, a multi-task deep learning model based on convolutional neural network is constructed to diagnose the multiple faults of machinery with heterogeneous information fusion.
Finally, a personalized federated learning framework is proposed to solve data heterogeneity across different machines using adaptive hierarchical aggregation strategy.
The case study on collected data from real machines verifies the effectiveness of the proposed framework. The result shows that the diagnosis accuracy could be improved significantly using the proposed personalized federated learning, especially for those machines with scarce fault samples. 

\end{abstract}

\section{Introduction}

Nowadays, with the increasing safety level of the whole society, major accidents caused by the failure of industrial equipment have attracted more and more attention. Especially in the fields related to petrochemical and energy, safety accidents caused by rotating machinery failure occur frequently. 
The failure and shutdown of rotating machinery will lead to the suspension of the whole production line and bring huge economic losses.
Intelligent fault diagnosis of rotating machinery can detect, identify and locate faults in time before they deteriorate. It is very important to ensure the safe and reliable operation of rotating machinery, and helpful to optimize maintenance strategy and spare parts.

Based on the collected vibration sensor data, many intelligent fault diagnosis methods have been proposed for rotating machinery using artificial intelligent algorithms, such as extreme learning machine \cite{ELM}, convolutional neural networks (CNN) \cite{CNN}, long short-term memory \cite{LSTM} and generative adversarial networks \cite{GAN}.
However, the above methods need to train an independent model for each machine and have poor generalization ability, since the data distribution varies across machines due to the difference in machine structure and working conditions. 
To address this problem, transfer learning and domain adaptation \cite{trans0} are introduced into fault diagnosis area.

However, most of the existing transfer learning methods can only deal with the variation of working conditions including rotating speed and load.  When applied to the fault diagnosis of multiple machines, these methods still need large amounts of fault samples covering all fault types to support the training process, which cannot be satisfied in many real industrial applications \cite{TL}.
In the actual industrial field, although there are many machines, most of them rarely have faults.
Meanwhile, the fault data of a machine is scarce and cannot cover all types of the faults that may happen. Therefore, it is difficult to accomplish model training or transfer learning with the data of a machine \cite{unb}.  
It is a feasible choice to use the data of different machines or even different factories to jointly build the model in a centralized data center. However, due to data privacy, many factories are reluctant to contribute the original sensor data, especially the fault samples, even if they are owned by the same entity. 
Meanwhile, the vibration data is sampled at high sampling frequency, and the data volume is large. Centralizing the data of multiple factories will consume a lot of IoT and data storage costs.

Federated learning, a novel paradigm to build AI models, can learn the knowledge from multiple participants while protecting the data privacy \cite{FL}. Federated learning allows multiple participants to train the model locally, and then aggregate the model updates without requiring participants to send the data to the central server \cite{fedproto}. However, the vanilla federated learning framework that builds a global model may not be effective for all participants due to the data heterogeneity \cite{FLBearing}, and thus, personalized federated learning with novel aggregation strategy \cite{PerFed} becomes a hot research topic.

Therefore, to address the problem of scarce fault samples and heterogeneous data across factories, this paper proposes a personalized federated learning framework for fault diagnosis of rotating machinery. Based on the similarity of time and frequency domain features extracted from the vibration signals, the rotating machines from multiple factories are categorized into machine groups using federated clustering algorithm. Then, a deep learning based fault diagnosis model composed of common feature exaction block and personalized multi-task classification block is constructed with the vibration features as inputs. Finally, a personalized federated learning framework is proposed through the adaptive aggregation of common feature extraction layer trained by all machines and the adaptive aggregation of personalized multi-task classification block for individual machine groups. 
The contributions of the paper are summarized as follows: 

\begin{itemize}
    \item A federated learning framework is proposed to deal with the scarce fault samples problem, which greatly improves the fault diagnosis accuracy. 
    \item Privacy-preserving federated clustering algorithm is proposed to categorize machines without seeing their data features.
    \item A personalized training method with adaptive hierarchical aggregation is proposed to solve data heterogeneity across different machines.
    \item A multi-task fault diagnosis model based on deep learning is constructed to diagnose multiple faults simultaneously by fusing the time and frequency features from vibration signals. 

\end{itemize}

The rest of this paper is organized as follows. Section 2 provides the review of machine fault diagnosis and personalized federated learning. Section 3 describes the detail process of the proposed personalized federated learning framework for machinery fault diagnosis. Section 4 verifies the proposed framework by a case study of actual machines from different factories. Section 5 concludes the paper.

\section{Related Work}
Fault diagnosis across the machines is a challenging work in real industrial scenarios due to the difference in data distribution. Therefore, transfer learning and domain adaptation methods are introduced to improve the performance of fault diagnosis models. 
For example, \cite{transfer} develops a new deep transfer learning based CNN to address the problem of transfer fault diagnosis under different working conditions. 
In \cite{trans2}, a deep adversarial transfer learning network is proposed for new emerging fault detection of target domain.
A distance metric named polynomial kernel induced maximum mean discrepancy is proposed in \cite{TL} to reuse diagnosis knowledge from one machine to the other. 

However, the above-mentioned methods still need complete training data of all fault types in both source domain and target domain, and cannot be guaranteed in practical implement. 
To reduce the dependence on data, a domain generalization framework Whitening-Net is proposed in \cite{transmachine} for the fault diagnosis across different machines and conditions based on causal mechanisms.

Although the above methods can obtain good results in transfer learning across machines, an independent model needs to be trained for each machine, which is inefficient in practical implementation. Federated learning uses the data from multiple participants while protecting the privacy, and is able to train a unified model using the knowledge of all participants.
\cite{FLBearing} proposes a fault diagnosis method based on federated learning and CNN, which allows different industrial participants to collaboratively train a global fault diagnosis model without sharing their local data.
However, the vanilla federated learning method cannot deal with data heterogeneity across the machines, as all participants share a single global model with the same model parameters. To address this problem, a clustered federated learning approach is presented in \cite{Clustered} to deal with distributions divergence of clients data by clustering clients based on cosine similarity. 
\cite{PerFed} proposes personalized federated learning framework in a cloud-edge architecture for intelligent IoT applications to cope with the heterogeneity issues in IoT environments.
Therefore, to address the current issues in practical machinery fault diagnosis, it is urgent to introduce personalized federated learning into the field of machinery fault diagnosis, and put forward a novel framework according to the characteristics of machinery faults. 

\section{Personalized Federated Learning Framework for Fault Diagnosis of Rotating Machinery}
In this section, firstly, the flowchart of the proposed clustering personalized federated learning framework is introduced. Then, the methods used in each part are described in detail.

\begin{figure*}[t]
    \centering
    \includegraphics[width=0.95\textwidth]{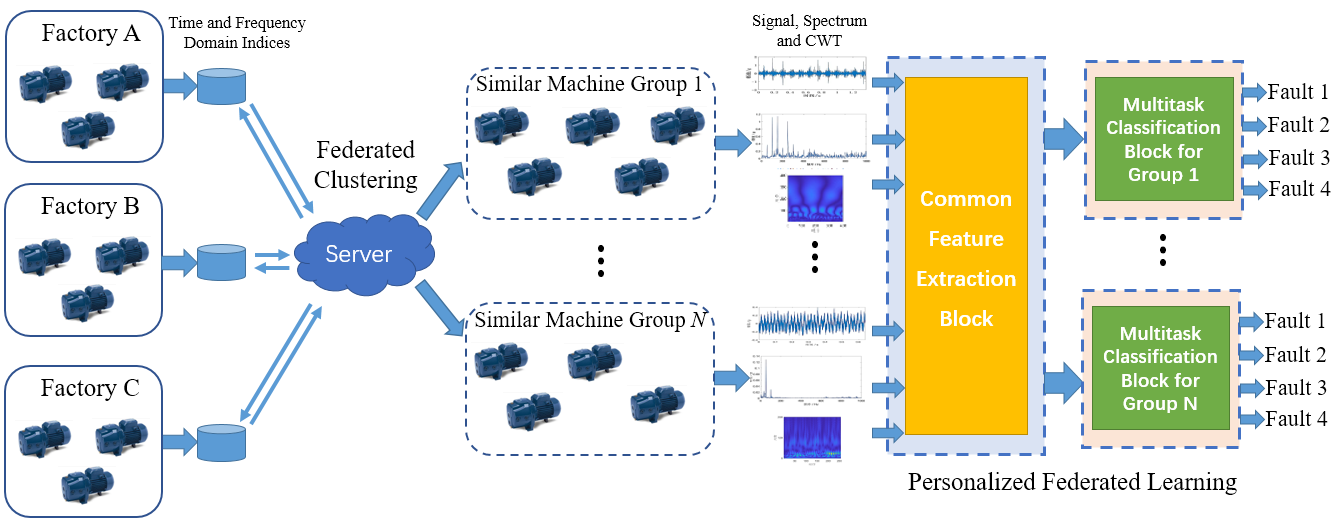} 
    \caption{Flowchart of the proposed clustering personalized federated learning framework.}
    \label{flowchart}
\end{figure*}

As shown in Figure \ref{flowchart}, based on the collected vibration data of machines, signal processing algorithms are carried out to extract the time frequency domain indices and features. Then, federated clustering is proceeded to categorize machines across factories into  machine groups based on the similarity of their time frequency indices. The personalized federated learning is carried out between similar machine groups across factories by adaptive hierarchical aggregation, which means the data of all machines are used to train the common feature extraction block, and the data of machines in each group are used to train the multi-task classification block for the similar machine group. Finally, in the online implementation, the faults of a machine can be diagnosed by the model composed of common feature extraction block and the multi-task classification block of its similar machine group.

\subsection{Data Collection and Preprocessing}
Vibration monitoring is one of the best ways to obtain the running state of rotating machinery \cite{VIB}. In this framework, multiple accelerometers are installed on the machine to collect the vibration information at different locations of the machine.
Then the time and frequency domain indices as listed in Table \ref{indices} are extracted from the original vibration signals for federated clustering.
In Table \ref{indices}, $k$X means the frequency related to $k$ multiplication of the rotating frequency.
Since the indices can reflect the overall vibration level and basic vibration components of the machine, it is reasonable to distinguish the similar machine groups based on the time and frequency domain indices.

\begin{table}[t]
\small
\small
    \centering
    \begin{tabular}{l|l}
        Time Domain Index & Frequency Domain Index \\\hline 
        Peak value & 0.5X amplitude \\
        Root mean square & 1X amplitude   \\
        Kurtosis & 2X amplitude    \\
        Skewness & 3X amplitude   \\
        Crest factor & 4X amplitude    \\ 
        Impulse factor & 5X amplitude    \\ 
    \end{tabular}
    \caption{Time and frequency domain indices.}
    \label{indices}
\end{table}

Moreover, the faults of machinery are mainly identified by the amplitudes of rotating frequency doubling and impact components in the vibration signals.
To make the deep learning model understand the fault features within vibration signals, and improve the generality of the deep learning model, Fourier transform and wavelet analysis are used to extract the frequency domain and time-frequency domain features of vibration signals \cite{CNN}. 

\begin{equation}\label{FFT}
    y(\omega )= \int_{-\infty }^{\infty }x(t)e^{-i\omega t}dt
  \end{equation}
    
\begin{equation}\label{wavelet2}
    C_{a}= \int x\left ( t \right )\bar{\Psi }_{a,b}\left ( t \right )dt
\end{equation}
\begin{equation}\label{wavelet3}
    \textbf{\textit{M}} = [C_{1},C_{2},\ldots,C_{S}]
\end{equation}
where $x(t)\in \mathbb{R} ^{T \times 1} $ stands for the original vibration signals in time domain, $y(\omega )\in \mathbb{R} ^{F \times 1}$ is the spectrum of $x$, $\textbf{\textit{M}} \in \mathbb{R} ^ {S \times T}$ is the continuous wavelet transform (CWT) of $x$. 
$\Psi_{a,b} \left ( t \right )$ is a continuous wavelet basis and $\bar{\Psi }_{a,b}\left ( t \right )$ is its complex conjugate in which scale parameter $a$ and translation parameter $b$ determine its shape and displacement, respectively. 
$C_{a}$ is the wavelet coefficients of signal $x(t)$ at the $a$-th scale. 
$T$ is the length of the signal, $F$ is the number of frequency and $S$ is the number of scales in wavelet transform.

\begin{figure}[!hbt]

    \centering
    \subfloat[Vibration signal]{\includegraphics[width=3.3in]{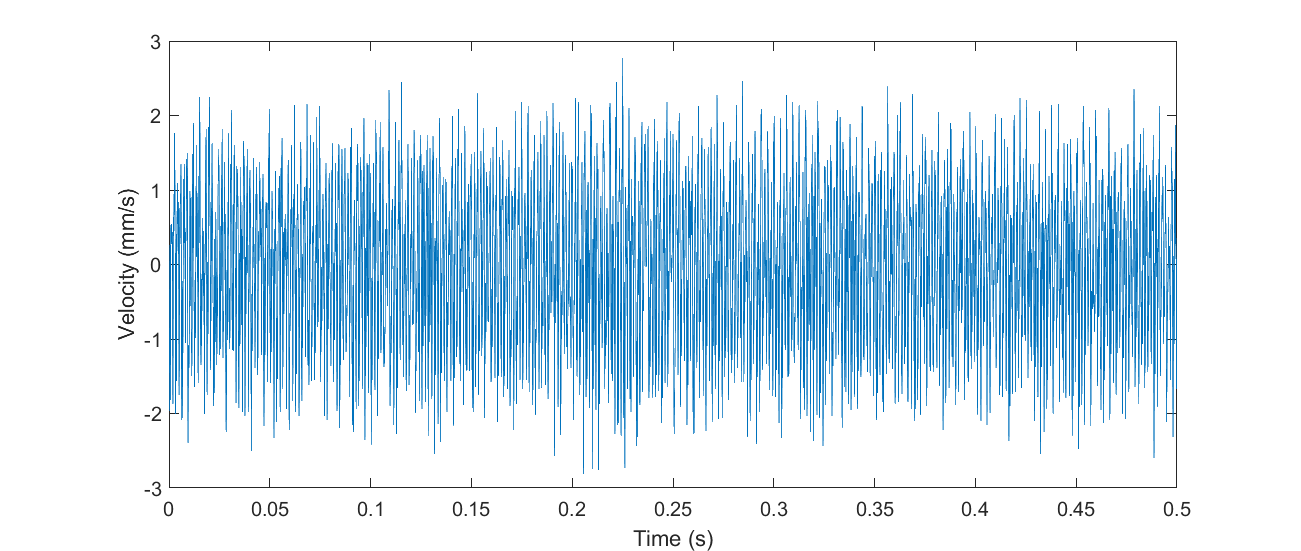}%
    \label{fig_first_case}}
    \hfil
    \subfloat[Spectrum]{\includegraphics[width=1.85in]{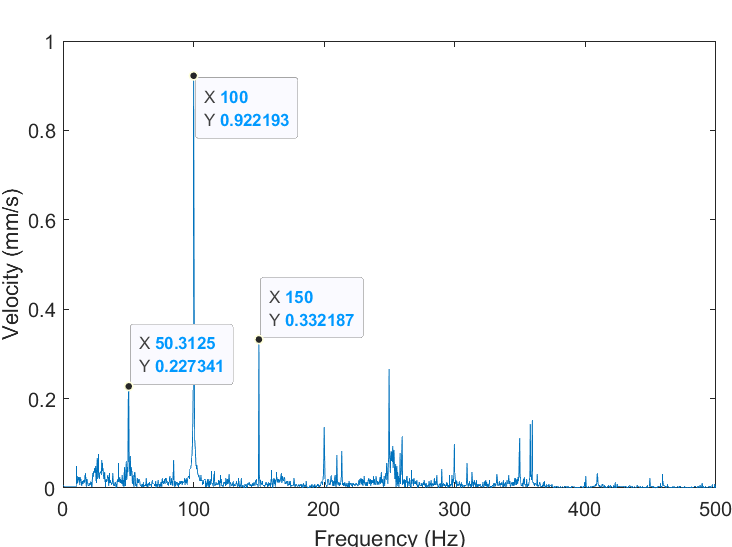}%
    \label{fig_second_case}}
    \subfloat[CWT]{\includegraphics[width=1.68in]{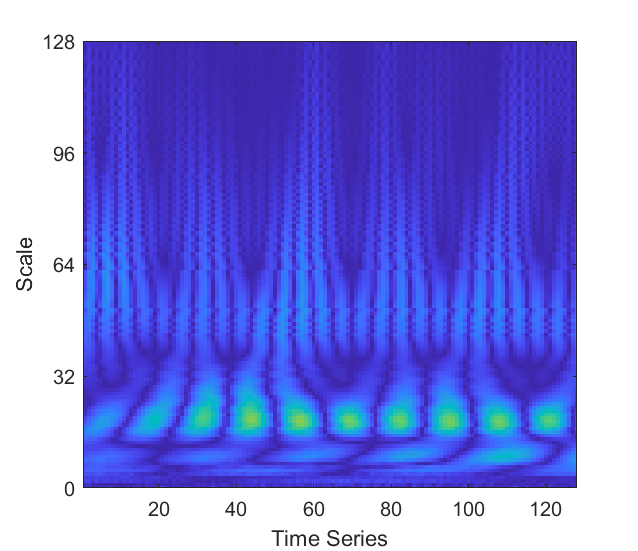}%
    \label{fig_third_case}}
    \caption{Time and frequency domain features of a vibration signal during a friction fault.}
    \label{tf_features}
\end{figure}

Figure \ref{tf_features} shows the original signal, spectrum and CWT of a velocity channel with a sampling frequency of 2.56 kHz during a misalignment fault of machine 9 in Table \ref{machines}. The rotating frequency of the machine is 50 Hz.
From the spectrum, the 1X, 2X and 3X amplitudes of the rotating frequency is obvious, which is the symptom of misalignment fault.
The CWT can show the non-stationary vibration characteristics in time-frequency domain, which is usually helpful in the diagnosis of bearing fault that generates intermittent impact components in the vibration signal.

\subsection{Multi-task Fault Diagnosis Model for Rotating Machinery}
 
Since CNN has been proved to be an effective model for processing vibration signals \cite{CNNreview}, a CNN-based fault diagnosis model for rotating machinery is constructed in this paper. Considering the domain knowledge of signal processing, spectrum is sensitive to the fault related to rotor, such as unbalance and misalignment, while the impact signal caused by bearing fault is obvious on the CWT.
Therefore, the original vibration signal, spectrum and CWT are used as the input of the fault diagnosis model simultaneously. Then a common feature extraction block is constructed to process multi-dimensional heterogeneous information input. At the same time, according to the fault types to be diagnosed, a multi-task classification block with multiple outputs is constructed for multiple faults.

\begin{figure}[!t]
    \centering
    \includegraphics[width=8.0cm]{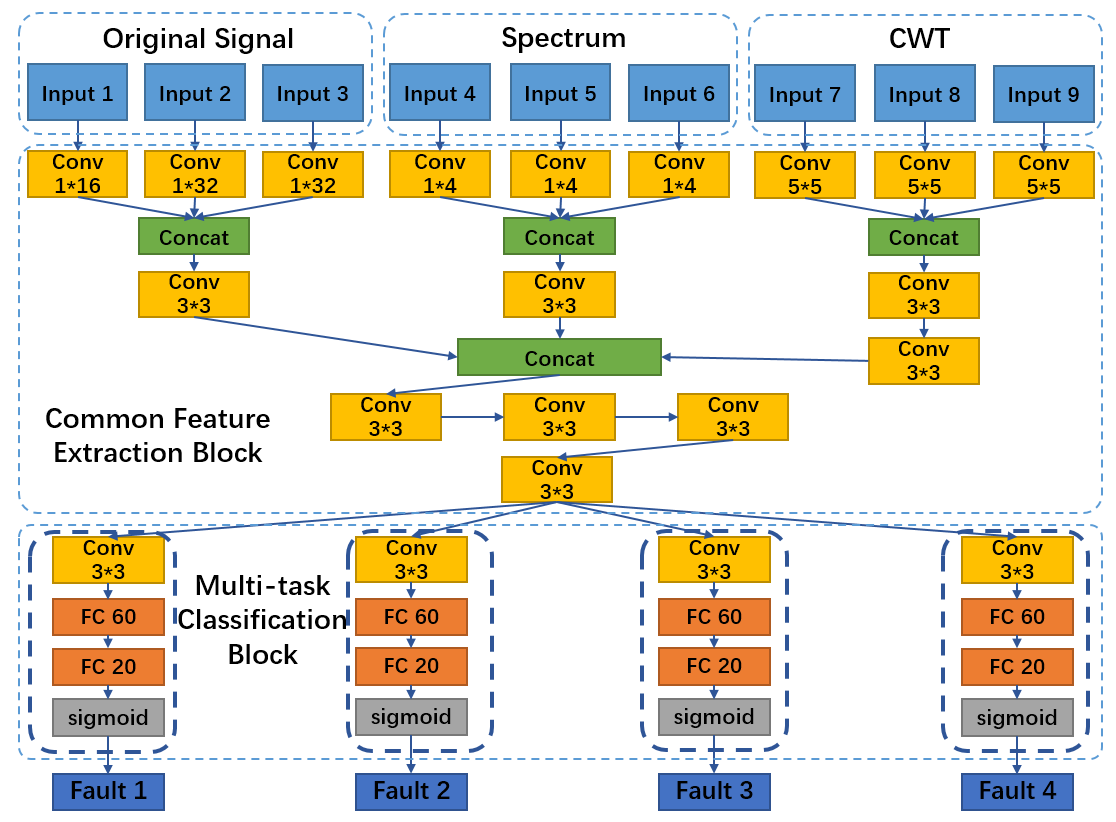} 
    \caption{Structure of the CNN-based multi-task fault diagnosis model for rotating machinery.}
    \label{CNNmodel}
\end{figure}

The detail structure of the fault diagnosis model is shown in Figure \ref{CNNmodel}. Conv represents the CNN block that consists of convolutional layer, activation layer and pooling layer. Vibration data of three channels in different sampling frequencies are used as the input of deep learning model. 
Input 1, input 2 and input 3 are original vibration signals of the three channels.
Input 4, input 5 and input 6 are spectrums of signals.
Input 7, input 8 and input 9 are CWT of signals.
Each type of data input is processed by two or three convolutional layers to obtain the feature maps in the same size. Then, the feature maps are concatenated for further feature extraction.
In the multi-task classification block, several fully connected batches are built to obtain the occurrence conditions of multiple faults, respectively.
In the case study of this paper, four fault types, i.e., unbalance, misalignment, bearing and friction, are considered.
The total number of parameters is about $1.4\times 10^{5}$, which is suitable for online deployment and real time inference.

Due to the serious imbalance between faults and normal samples, during the local training of the multi-task fault diagnosis model,
an adaptive sensitive cost method is proposed to accelerate and balance the training process.
\begin{align}\label{adpativecons}
    \begin{split}
        T_{i}&=(\sum_{1}^{N}f_{i})/f_{i} \\
        C_{i}^{j}&= (l_{i}^{j} (1-r_{i})+1)\times T_{i}\\
        L_{total}&=\sum_{i=1}^{N}\sum_{j=1}^{n}(y_{i}^{j}-l_{i}^{j})^2 C_{i}^{j}
    \end{split}
\end{align}
where $T_{i}$ is the sensitive coefficient of the $i$-th fault diagnosis task, $f_{i}$ is the $F_{1}$ score of the $i$-th task in the last training step and $N$ is the number of fault types. $r_{i}$ represents the rate of data of the $i$-th fault in the training dataset. $l_{i}^{j}$ represents the label of the $j$-th sample for the $i$-th fault in the training dataset. $l_{i}^{j}=1$ means the occurrence of the fault, while $l_{i}^{j}=0$ means none fault. 
$C_{i}^{j}$ represents the coefficient for the $i$-th fault of the $j$-th sample. Thus, the fault samples will obtain a higher weight than the non-fault samples, which is used to alleviate  the problem of scarce fault samples. 
$y_{i}^{j}$ is the output of the $j$-th sample for the $i$-th fault, $n$ is the number of data samples, and $L_{total}$ stands for the loss function to be minimized in the training process.
By this means, the fault with lower $F_{1}$ score will be given a larger weight in the loss function to accelerate its training. 
The $F_{1}$ score is calculated by \cite{F1}:
\begin{align}\label{f1scale}
    \begin{split}
        Accuracy&=\frac{TP+TN}{TP+TN+FP+FN}\\
        Recall&=\frac{TP}{TP+FN}\\
        Precision&=\frac{TP}{TP+FP}\\
        F_{1}&=\frac{2\cdot Precision \cdot Recall  }{Precision +Recall }
    \end{split}
\end{align}
where $TP$ is the true positive rate, $TN$ is the true negative rate, $FP$ is the false positive rate and $FN$ is the false negative rate.


\subsection{Personalized Federated Learning with Adaptive Aggregation}

In order to solve the problem of scarce fault samples and data heterogeneity, a personalized federated learning method based on hierarchical aggregation is proposed based on the clustering results of similar machine groups.

\subsubsection{Federated Clustering}
\begin{figure}[t]
    \centering
    \includegraphics[width=7cm]{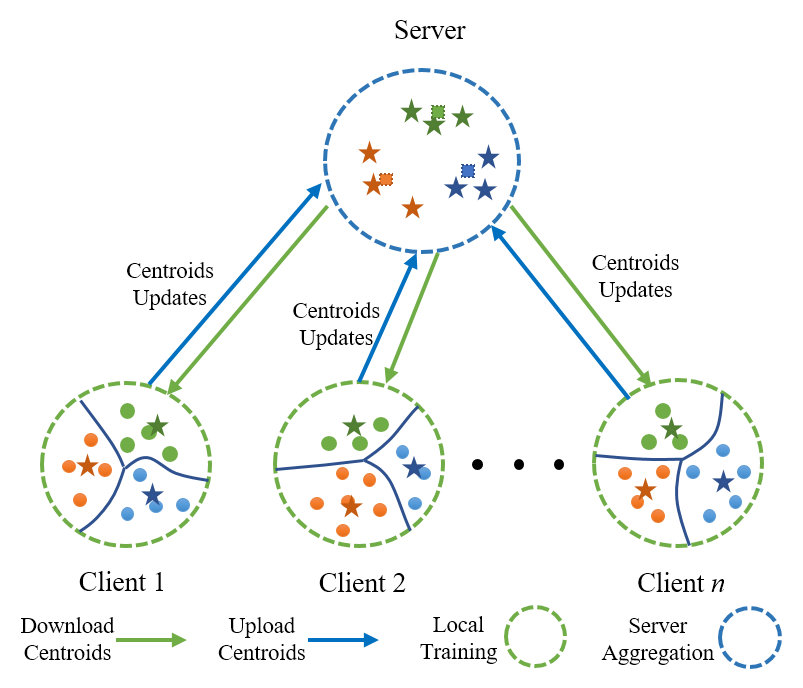} 
    \caption{Federated $k$-means for machine clustering.}
    \label{fedclustering}
\end{figure}

Due to the different distribution of data characteristics among a number of machines, it is difficult to train a good federated learning model using the data of all machines \cite{Clustered}. At the same time, due to the large number of rotating machines in the factory, it is unrealistic to train or transfer a model for each machine. Therefore, in the proposed framework, based on the characteristics of vibration data, multiple machines are divided into multiple similar machine groups. Machines belonging to the same similar machine group share the parameters of the diagnostic model. Moreover, due to the problem of data privacy among different factories, this paper adopts a clustering method based on federated $k$-means \cite{fedclus}. The specific steps are illustrated as in Figure \ref{fedclustering}.

The time and frequency domain indices listed in Table \ref{indices}, which is represented by small and solid circle in Figure \ref{fedclustering}, are used as the data for federated clustering. In order to avoid the change of machine data distribution caused by faults, only the normal data of each machine is used for clustering similar machine groups. 
Each client represents a factory participating in the federated clustering. During the training process of federated $k$-means, the clients carry out the $k$-means training locally and upload the cluster centroids, which is represented by the stars in Figure \ref{fedclustering}, to the server. Then, the server receives the cluster centroids from clients and aggregates the cluster centroids to obtain the global cluster centroids represented by the square in the figure. Next, the clients download the global cluster centroids from the server and continue the local training. Finally, the federated clustering is converged after multiple iterations when the gaps between all the local cluster centroids and the corresponding global cluster centroids are smaller than pre-defined threshold. The machine is  categorized to a group if its majority data belongs to the group. 

\subsubsection{Personalized Federated Learning}
Since the machine fault mechanism and basic vibration characteristics are consistent across the rotating machinery, the common feature extraction block of the fault diagnosis model is expected to extract the common vibration features from all machines.
Then, considering the difference in machine parameters and operating conditions, the distribution of the common vibration features may be different across the machines.
Each similar machine group that contains machine in similar time domain features is designed to have its own multi-task classification block. The common feature layer is proposed to cooperate with the multi-task classification block corresponding to its own similar machine group.

\begin{figure}[t]
    \centering
    \includegraphics[width=8.4cm]{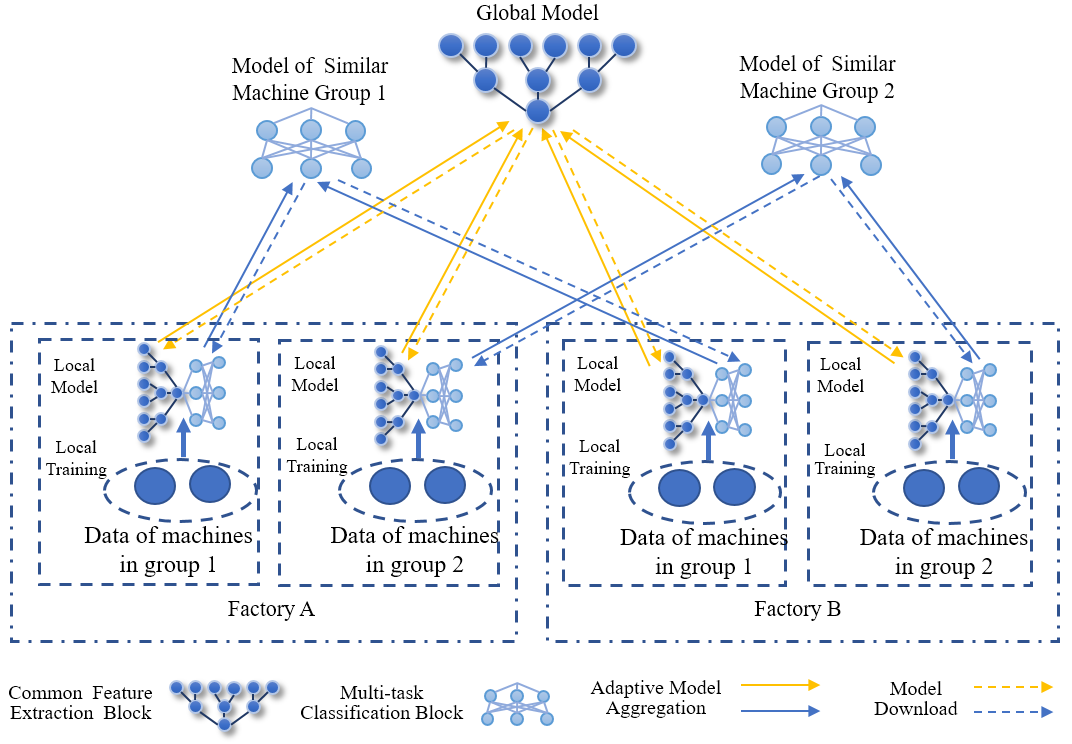} 
    \caption{Training and aggregation process of personalized federated learning.}
    \label{perFedLearning}
\end{figure}

Figure \ref{perFedLearning} shows the training and aggregation process of the proposed personalized federated learning method. It is assumed that Factory A and Factory B are the two factories participating in federated learning. 
Supposing all the machines are divided into two similar machine groups. For each factory, two agents are built, corresponding to two similar machine groups.
For each agent of each factory, the data of the machines corresponding to the same similar machine group in the factory are summarized to training the same local model.
Compared with training a local model for the data of each machine, this method reduces the transmission of model parameters in the network, especially in the case of a large number of machines.
For agents belonging to the same similar machine group in all factories, model aggregation is carried out by a variant of FedAvg \cite{fedavg} (as described in next subsection). Finally, two fault diagnosis models are obtained for two similar machine groups, respectively.

\subsubsection{Adaptive Aggregation}
Since the data of the machines in each factory has different numbers of data samples especially the number of fault samples, the training speed of each agent varies greatly. If the aggregation rates are equal among the agents (e.g., FedAvg aggregation strategy \cite{fedavg}), the local model trained with more fault data will have a larger impact on the global model, which will result in a slow training of the local model trained by scarce fault data.
Therefore, in this paper, an adaptive aggregation strategy is proposed to balance the training of all agents by adjusting aggregation rates according to the $F_{1}$ scores of all agents.
Algorithm \ref{alg:algorithm} shows the detail training process with adaptive aggregation rates of personalized federated learning.

\begin{algorithm}[!t]
    \caption{Training process of personalized federated learning with adaptive aggregation rate.}
    \label{alg:algorithm}
    \textbf{Input}: Model, learning parameters, training set\\
    \textbf{Output}: $\mathbf{w}_{global}, \mathbf{\theta}_{group}^{1}, \mathbf{\theta}_{group}^{2},..., \mathbf{\theta}_{group}^{N_{group}}$\\
    \begin{algorithmic}[1] 
 
    \STATE \textbf{(1). Industrial Participant side:}
    \FOR {$i=1,2,..,N_{factory}$}
        \STATE Initialize $N_{group}$ federated learning agents.
    \ENDFOR
 
   \FOR [run in parallel] {$i=1,2,..,N_{factory}$} 
        \FOR [run in parallel] {$j=1,2,..,N_{group}$} 
            \FOR {local training round $t=1,2,...,M$}
                \STATE $\mathbf{w}_{i}^{j}\leftarrow \mathbf{w}_{global}$ \COMMENT{Load the global weight}
                \STATE $\mathbf{\theta}_{i}^{j}\leftarrow \mathbf{\theta}_{group}^{j}$
                \COMMENT{Load the group weight}
                \STATE Train the local model 
                using the data of the machines of factory $i$ in group $j$.  
                \STATE Upload $\mathbf{w}_{i}^{j},\mathbf{\theta}_{i}^{j} $ and $F_{i}^{j}$ to the cloud server.
            \ENDFOR
        \ENDFOR
    \ENDFOR

    \STATE \textbf{(2). Cloud Server side:}
    \STATE Receive  $\mathbf{w}_{i}^{j},\mathbf{\theta}_{i}^{j} $ and $F_{i}^{j}$ from each participant. 
    \STATE $c_{1}^{1},...,c_{N_{factory}}^{ N_{group}} \leftarrow$ adaptive($F_{1}^{1},...,F_{N_{factory}}^{ N_{group}}$) 
     \COMMENT{Update aggregation weights for global model }
     
    \STATE $\mathbf{w}_{global} \leftarrow$ aggregate($\mathbf{w}_{1}^{1},...,\mathbf{w}_{i}^{j},c_{1}^{1},...,c_{N_{factory}}^{ N_{group}}$) 
    \FOR [Update aggregation weights for group model]{$j=1,2,...,N_{group}$}
        \STATE $d_{1}^{j},...,d_{N_{factory}}^{ j} \leftarrow$ adaptive($F_{1}^{j},...,F_{N_{factory}}^{ j}$) 

        \STATE $\mathbf{\theta}_{group}^{j} \leftarrow$ aggregate($\mathbf{\theta}_{1}^{j},...,\mathbf{\theta}_{N_{factory}}^{j},d_{1}^{j},...,d_{N_{factory}}^{ j}$) 
    \ENDFOR
    \STATE \textbf{return} $\mathbf{w}_{global}, \mathbf{\theta}_{group}^{1}, \mathbf{\theta}_{group}^{2},..., \mathbf{\theta}_{group}^{N_{group}}$
    \end{algorithmic}
\end{algorithm}

In Algorithm \ref{alg:algorithm}, $N_{factory}$ stands for the number of factories participating the federated learning and $N_{group}$ stands for the number of similar machine groups.  $\mathbf{w}$ represents the parameters of common feature extraction block and $\theta$ represents the parameters of multi-task classification block.

After each training epoch, based on the accuracy of training data of each client, the $F_{1}$ score of all faults are calculated and averaged.
Then, the local model and average $F_{1}$ score of each client are uploaded to the server.
The server allocates model aggregation weights according to $F_{1}$ scores of different clients. If the $F_{1}$ score of a client is high, its aggregation weight will be low. On the contrary, if $F_{1}$ score is low, the client will get high aggregation weight to speed up its training. The adaptive function is identified by:
\begin{equation}\label{aggweight}
    c_{k}= \sum_{m} F_{m} / F_{k}^{2} 
\end{equation}
where $F_{m}$ is the $m$-th $F_{1}$ score input of the adaptive function, $c_{k}$ and $F_{k}$ are the output aggregation weight and $F_{1}$ score of the $k$-th client, respectively. 

Next, the server aggregates the common feature extraction block of all the local models as the global common feature extraction block. The multi-task classification blocks of the local models corresponding to the same similar machine group across factories are aggregated as the multi-task classification block of the group. 
The aggregate function is identified by:
\begin{equation}\label{agg}
    \alpha =\sum_{k} \alpha_{k} c_{k}/ \sum_{k} c_{k}
\end{equation}
where $\alpha_{k}$ represents the $k$-th input network parameters, $c_{k}$ is the corresponding aggregation weight, $\alpha$ represents the aggregated network parameters.
The adaptive aggregation is carried out in both global model aggregation and group model aggregation.

Finally, the personalized federated learning achieves convergence when the $F_{1}$ scores of all local models stop increasing.

\subsection{Model Deployment}
After obtaining the complete personalized federated learning model, participating factories can use model to diagnose the fault condition of machines.
There are three typical application scenarios of the personalized federated learning model:

\begin{itemize}
    \item For a machine included in the training process, the real-time collected data are sent into the fault diagnosis model that is consist of the global common feature extraction block and the multi-task classification block of its similar machine group for fault diagnosis.
    \item For a new machine installed in a factory, with the centroids obtained in federated clustering, the new machine can be classified into a certain similar machine group. Then, using the complete model of the group, the future fault condition can be diagnosed online. 
    \item For a factory which was not a participant of the federated learning, each machine in the factory can get its similar machine group information with the centroids obtained from the server. Then, using the complete model of the group, the fault condition of the machines can be diagnosed.  

\end{itemize}

Then the business value obtained from the diagnosis results should be returned to each factory participating in the federated learning, according to its contribution and incentive mechanism of the entire ecosystem, which is out of scope of this paper.

\section{Case Study}
To verify the effectiveness of the proposed personalized federated learning framework, a case study based on the data collected from factories is carried out. The fault diagnosis results of the proposed framework are analyzed and further compared with other methods.

In this case study, 13 machines from different factories, named as A, B and C, are selected as data source. 
In the data collection, accelerometers are installed on the surface of the driven or non-driven end of the machines. Each accelerometer has three channels: acceleration under sampling rate of 51.2 kHz, acceleration under sampling rate of 5.12 kHz and velocity obtained by integration under sampling rate of 2.56 kHz.

The detail information of the machines and data are listed in Table \ref{machines}. The selected machines are pumps in different types and rated power. All the machines have different numbers of data samples and fault types, while none of the machines has the complete fault types. Fault type 1 to 4 represent unbalance, misalignment, bearing and friction, respectively. The fault rate in the followed bracket shows the ratio of the fault samples corresponding to each fault type in the total data of the machine.

\begin{table*}[t]
\small
    \centering
    \begin{tabular}{l|l|l|l|l|l}
        No. & Factory & Type & Rated Power & Data Samples & Fault Type (Rate)\\\hline 
        1 & A & Magnetic pump & 45kW & 1271 & 3 (2.8\%), 4 (21.7\%)\\
        2 & & Magnetic pump & 45kW & 2416 & 4 (82.9\%)\\
        3 & & Centrifugal pump & 280kW & 300 &  \\
        4 & &  Centrifugal pump & 280kW & 1144 &  3 (11.5\%) \\
        5 & &  Centrifugal pump & 280kW & 1020 & 3 (3.5\%) \\ \hline 
        6 & B& Centrifugal pump & 90kW & 1064 & 4 (51.9\%) \\  
        7 & &  Centrifugal pump & 90kW & 712 & 4 (5.1\%) \\   
        8 & &  Magnetic pump & 45kW & 824 & 3 (1.5\%), 4 (24.8\%)\\
        9 & &  Centrifugal pump & 280kW & 1848 & 1 (4.5\%), 2 (45.5\%), 3 (12.3\%) \\ 
        10 & & Centrifugal pump & 280kW & 1060 & 3 (2.3\%) \\ \hline 
        11 & C &  Centrifugal pump & 90kW & 872 & 4 (31.7\%) \\  
        12 & &  Centrifugal pump & 90kW & 1018 & 2 (5.9\%), 3 (6.5\%), 4 (37.1\%) \\   
        13 & &  Centrifugal pump & 280kW & 2360 & 1 (32.5\%), 3 (48.8\%) \\ 
    \end{tabular}
    \caption{Information of machines.}
    \label{machines}
\end{table*}

At data preprocessing stage, the time and frequency domain indices, spectrum and CWT of the three channels are calculated based on the vibration signals. The size of the signal and three types of features are listed in Table \ref{featuresize}. 
The indices of three channels are combined as the data source of federated clustering,  
and the other three types of data are used as the input of fault diagnosis model.

\begin{table}[t]
\small
    \centering
    \begin{tabular}{l|l|l}
        Signal Channel & Feature Type & Size \\\hline 
        8 kHz velocity & Original signal & $4096 \times 1$\\
                    & Indices &  $10 \times 1$  \\
                    & Spectrum & $128 \times 1$   \\
                    & CWT & $128 \times 128$   \\\hline 
        16 kHz acceleration & Original signal & $8192 \times 1$  \\
                    & Indices & $10 \times 1$   \\
                    & Spectrum & $128 \times 1$   \\
                    & CWT & $256 \times 256$  \\\hline 
        32 kHz acceleration & Original signal & $16384 \times 1$ \\
                    & Indices & $10 \times 1$   \\
                    & Spectrum & $256 \times 1$    \\
                    & CWT & $384 \times 384$   \\
    \end{tabular}
    \caption{Size of data features.}
    \label{featuresize}
\end{table}

With the indices as input, the result of federated clustering is listed in Table \ref{Clustering}. 
It can be seen that the machines with similar power are grouped, exceeding the impact of machine type. The last column lists the identifier of the model that is trained by the data of the machine in the same row.

\begin{table}[H]
\small
    \centering
    \begin{tabular}{l|l|l}
        Similar Machine  & Factory:  & Local \\
        Group & Machine No. & Model No.\\\hline 
        1 & Factory A:  1, 2 & A1\\
                    & Factory B: 6, 7, 8 & B1\\
                    & Factory C: 11, 12& C1\\ \hline 
        2 & Factory A: 3, 4, 5 & A2\\
                    & Factory B: 9, 10 & B2\\
                    & Factory C: 13 & C2\\ 
    \end{tabular}
    \caption{Result of federated clustering.}
    \label{Clustering}
\end{table}


During the training of the personalized federated framework, two agents are created in each factory corresponding to the two similar machine groups. The data of the machines in each group of each factory are summarized in the training process, e.g. the data of machine 1 and machine 2 are summarized in factory A for the training of the same local model. 
Then the local models of the 6 agents are aggregated hierarchically with the adaptive aggregation rates. 
Specifically, the common feature extraction blocks of all 6 local models are aggregated to obtain the global common feature extraction block. The multi-task classification blocks of model A1, B1 and C1 are aggregated to obtain the multi-task classification block for similar machine group 1, and the multi-task classification blocks of model A2, B2 and C2 are aggregated to obtain the multi-task classification block for similar machine group 2.

The training of the personalized federated learning framework is carried out on a server with a P1000 GPU. The federated learning sever and agents are simulated by multi-threading in \textit{Python} environment. The deep learning model based on CNN is realized by \textit{Tensorflow}.
The data of all machines are divided into five datasets for 5-fold cross validation, which means each training experiment takes four datasets as training data and the rest dataset as test data. The training of each combination of datasets takes about 1000 epochs to obtain convergence in about eight hours.
Table \ref{PFLresult} shows the fault diagnosis result including average $F_{1}$ score, accuracy, precision and recall for all faults of all machines obtained by the personalized federated learning framework.
The results of machine 3 are not list in the paper, as it has no fault samples.

\begin{table}[t]
\small
    \centering
    \begin{tabular}{l|l|l|l|l|l}
        Machine  & Fault & $F_{1}$ & Accuracy & Precision & Recall \\ 
        No. & Type  &  & & & \\\hline  
        1 & 3 & 0.526 & 0.965 & 0.417 & 0.714 \\
          & 4 & 1.0 & 1.0 & 1.0 & 1.0 \\\hline
        2 & 4 & 0.977 & 0.961 & 0.955 & 1.0 \\\hline
        4 & 3 & 0.962 & 0.991 & 0.926 & 1.0 \\\hline
        5 & 3 & 1.0 & 1.0 & 1.0 & 1.0 \\\hline
        6 & 4 & 0.939 & 0.934 & 0.908 & 0.973 \\\hline
        7 & 4 & 1.0 & 1.0 & 1.0 & 1.0 \\\hline
        8 & 3 & 1.0 & 1.0 & 1.0 & 1.0 \\
          & 4 & 0.95 & 0.976 & 0.905 & 1.0 \\\hline
        9 & 1 & 0.561 & 0.932 & 0.390 & 1.0\\
          & 2 & 0.917 & 0.924 & 0.912 & 0.923\\
          & 3 & 0.688 & 0.921 & 0.681 & 0.696\\\hline
        10& 3 & 0.889 & 0.995 & 0.8 & 1.0 \\\hline
        11& 4 & 0.774 & 0.839 & 0.696 & 0.873\\\hline
        12& 2 & 0.923 & 0.990 & 0.857 & 1.0\\
          & 3 & 0.667 & 0.951 & 0.556 & 0.833\\
          & 4 & 0.960 & 0.970 & 0.960 & 0.960\\\hline
        13& 1 & 0.890 & 0.932 & 0.929 & 0.855\\
          & 3 & 0.843 & 0.824 & 0.748 & 0.965\\

    \end{tabular}
    \caption{Fault diagnosis result of the personalized federated framework.}
    \label{PFLresult}
\end{table}

From Table \ref{PFLresult}, it can be seen that all the machines obtain accurate fault diagnosis results for all fault after personalized federated learning, even when the fault rate is very low, such as the fault 4 of machine 12. The machine with more fault data obtains higher diagnosis accuracy. The recall rates of all machines and fault types remain high, which is very helpful to find the occurrence of fault in real time. 

To demonstrate the superiority of the proposed framework, two common methods for fault diagnosis of multiple rotating machines are used to compared with the proposed framework. Single machine method builds a fault diagnosis model for each machine using its own data only. Vanilla federated learning (Vanilla FL) method aggregates the local model of all machines as a single global model. Moreover, clustering FL, which is implemented by training a model for each similar machine group, is also proposed in this paper for comparison. 

\begin{table}[t]
\small
    \centering
    \begin{tabular}{l|l|l|l|l|l}
        Machine  & Fault & Single  & Vanilla  & Clustering   & Personalized   \\ 
        No. & Type  &  Machine &  FL & FL & FL   \\\hline  
        1 & 3 &  \gray{\textbf{0.516}} & 0.5 & 0.5 & \textbf{0.526}   \\
          & 4 &  \textbf{1.0} & \textbf{1.0} &  \textbf{1.0} & \textbf{1.0} \\\hline
        2 & 4 & \gray{\textbf{0.972}} & 0.970 & 0.970 &  \textbf{0.977}  \\\hline
        4 & 3 & \textbf{0.964} & 0.935 & 0.915 &  \gray{\textbf{0.962}}  \\\hline
        5 & 3 & 0.875 & \gray{\textbf{0.933}} & 0.824 &  \textbf{1.0}  \\\hline
        6 & 4 & \textbf{0.943} & 0.935 & \textbf{0.943} &  0.939  \\\hline
        7 & 4 & \gray{\textbf{0.933}} & 0.574 & 0.794 &  \textbf{1.0}  \\\hline
        8 & 3 & \gray{\textbf{0.8}} & 0.667 & \gray{\textbf{0.8}} & \textbf{1.0}  \\
          & 4 & \textbf{1.0} & 0.943 & \gray{\textbf{0.976}} &   0.95 \\\hline
        9 & 1 & 0.548 & 0.557 & \textbf{0.708} &  \gray{\textbf{0.561}}  \\
          & 2 & \textbf{0.977} & 0.904 & \gray{\textbf{0.96}} &  0.917  \\
          & 3 & \textbf{0.891} & 0.8 & \gray{\textbf{0.878}} &  0.688  \\\hline
        10& 3 & 0.046 & 0.769 & \gray{\textbf{0.833}} & \textbf{0.889}  \\\hline
        11& 4 & 0.730 & 0.660 & \textbf{0.781} & \gray{\textbf{0.774} } \\\hline
        12& 2 & 0.786 & \gray{\textbf{0.815}} & 0.8 &  \textbf{0.923}  \\
          & 3 & 0.265 & \gray{\textbf{0.634}} & 0.579 & \textbf{0.667}  \\
          & 4 & \textbf{0.973} & \textbf{0.973} & \textbf{0.973} &  0.960  \\\hline
        13& 1 & \textbf{0.965} & 0.942 & \gray{\textbf{0.962}} &  0.890  \\
          & 3 & \gray{\textbf{0.929}} & 0.927 & \textbf{0.974} & 0.843   \\\hline
        Average & & 0.818 & 0.831 & \textbf{0.868} & \textbf{0.855}\\

    \end{tabular}
    \caption{Detail $F_{1}$ scores in the comparison of the three fault diagnosis methods.}
    \label{Compareresult}
\end{table}

The detail $F_{1}$ scores of diagnosis result of all machines in all fault types are listed in Table \ref{Compareresult}.
It can be seen that the personalized FL framework obtains higher diagnosis accuracies than the other two methods for most machines and faults. The vanilla FL method fails to improve the diagnosis accuracy than single machine models, since it cannot deal with the data heterogeneity. Considering the data heterogeneity, personalized FL trains personalized fault diagnosis models for machines by the means of federated clustering and hierarchical aggregation, which  improves the generality of the models. Therefore, personalized FL improves the diagnosis accuracies greatly when the fault rates of the training data are low, e.g. fault 1 of machine 9 and fault 3 of machine 12, and maintains the diagnosis accuracies when the fault data in the training set of a machine is enough to train a good model.

From the statistical results of $F_{1}$ scores by fault types shown in Table \ref{CompareFault}, we can see that the two methods proposed in this paper have the highest diagnosis accuracy on the two fault types respectively, which verifies that similar machine clustering can relieve the data heterogeneity across machines.
Table \ref{FaultRate} shows the diagnosis result under different fault rates in the dataset. It can be seen that when there are few fault samples, all federated learning methods can improve the diagnosis accuracy and personalized FL outperforms both vanilla and clustering FL. The improvement of $F_{1}$ scores between the single machine model and personalized FL model is the highest when fault rate is less than 5\%. Meanwhile, the improvement decreases while the fault rate increasing.  When the fault samples are sufficient, the single machine model might be good enough and the improvements of all FL methods become marginal. In other words, federated learning is beneficial for the factories having machines with scarce fault samples, which is very common especially for the newly installed machines.

\begin{table}[t]
\small
    \centering
    \begin{tabular}{l|l|l|l|l}
        Fault & Single  & Vanilla    & Clustering  & Personalized   \\ 
        Type  &  Machine &  FL & FL  & FL  \\\hline  
        1 &  0.782 & 0.773 & \textbf{0.851} & 0.746  \\
        2 &  0.909  & 0.872 & 0.903 & \textbf{0.919} \\
        3 &  0.708 & 0.790 & \textbf{0.813} &  0.806  \\
        4 &  0.946 & 0.900 & 0.937 & \textbf{0.951} \\
        
    \end{tabular}
    \caption{The average $F_{1}$ scores for all fault types using different fault diagnosis methods.}
    \label{CompareFault}
\end{table}

\begin{table}[t]
\small
    \centering
    \begin{tabular}{l|l|l|l|l|l}
        Fault &  Fault & Single  & Vanilla    & Clustering  & Personal-   \\ 
        Rate  &  Number &  Machine &  FL & FL  &ized FL  \\\hline  
        0 - 5\% & 5 & 0.543 & 0.661 & 0.718 & \textbf{0.746}   \\
        5 - 15\% & 5 & 0.781 & 0.772 & 0.808 & \textbf{0.819}   \\
        15 - 30\% & 2 &  \textbf{1.0}  & 0.977 & 0.991 &  0.980 \\
        30 - 50\% & 5 & 0.934 & 0.904 & \textbf{0.948} &  0.879  \\
        $>$50\%  & 2 & 0.963 & 0.959 & 0.962 & \textbf{0.965} \\
    \end{tabular}
    \caption{The average $F_{1}$ scores under different fault rates using different fault diagnosis methods.}
    \label{FaultRate}
\end{table}

\section{Conclusion and Future Work}
In this paper, a personalized federated learning framework is proposed for multi-task fault diagnosis of rotating machinery. The machines from different factories are clustered into similar machine groups according to the features of their vibration signals without compromising data privacy. Based on the multi-task fault diagnosis model constructed, the personalized federated learning is carried out by aggregating the common feature extraction block of all local model of machines and aggregating the multi-task classification block in each similar machine group.
The case study on the machines from different factories shows the advantages of the proposed clustering and personalized federated learning methods, over the single machine model and vanilla federated learning methods in terms of higher diagnosis accuracy for all types of failures investigated.

In the future, incentive mechanism will be further studied to encourage factories joining personalized federated learning using monetary of non-monetary schemes, and thus to build a prosperous ecosystem having sufficient machines with similar data and fault samples for achieving higher diagnosis accuracy and the corresponding business value.

\bibliography{aaai22}

\begin{thebibliography}{20}
\providecommand{\natexlab}[1]{#1}

\bibitem[{Cao et~al.(2019)Cao, Qian, Zareipour, Huang, and Zhang}]{LSTM}
Cao, L.; Qian, Z.; Zareipour, H.; Huang, Z.; and Zhang, F. 2019.
\newblock Fault Diagnosis of Wind Turbine Gearbox Based on Deep Bi-Directional
  Long Short-Term Memory Under Time-Varying Non-Stationary Operating
  Conditions.
\newblock \emph{IEEE Access}, 7: 155219--155228.

\bibitem[{Guo et~al.(2019)Guo, Lei, Xing, Yan, and Li}]{trans0}
Guo, L.; Lei, Y.; Xing, S.; Yan, T.; and Li, N. 2019.
\newblock Deep Convolutional Transfer Learning Network: A New Method for
  Intelligent Fault Diagnosis of Machines With Unlabeled Data.
\newblock \emph{IEEE Transactions on Industrial Electronics}, 66(9):
  7316--7325.

\bibitem[{Guo et~al.(2020)Guo, Zhang, Yang, Lyu, and Gao}]{CNN}
Guo, S.; Zhang, B.; Yang, T.; Lyu, D.; and Gao, W. 2020.
\newblock Multitask Convolutional Neural Network With Information Fusion for
  Bearing Fault Diagnosis and Localization.
\newblock \emph{IEEE Transactions on Industrial Electronics}, 67(9):
  8005--8015.

\bibitem[{Hang, Yang, and Xing(2019)}]{unb}
Hang, Q.; Yang, J.; and Xing, L. 2019.
\newblock Diagnosis of Rolling Bearing Based on Classification for High
  Dimensional Unbalanced Data.
\newblock \emph{IEEE Access}, 7: 79159--79172.

\bibitem[{Isham et~al.(2019)Isham, Leong, Lim, and Bin~Ahmad}]{ELM}
Isham, M.~F.; Leong, M.~S.; Lim, M.~H.; and Bin~Ahmad, Z.~A. 2019.
\newblock Intelligent wind turbine gearbox diagnosis using VMDEA and ELM.
\newblock \emph{Wind Energy}, 22(6): 813--833.

\bibitem[{Konečný et~al.(2017)Konečný, McMahan, Yu, Richtárik, Suresh, and
  Bacon}]{FL}
Konečný, J.; McMahan, H.~B.; Yu, F.~X.; Richtárik, P.; Suresh, A.~T.; and
  Bacon, D. 2017.
\newblock Federated Learning: Strategies for Improving Communication
  Efficiency.
\newblock arXiv:1610.05492.

\bibitem[{Li et~al.(2020)Li, Huang, He, Wang, Li, and Li}]{trans2}
Li, J.; Huang, R.; He, G.; Wang, S.; Li, G.; and Li, W. 2020.
\newblock A Deep Adversarial Transfer Learning Network for Machinery Emerging
  Fault Detection.
\newblock \emph{IEEE Sensors Journal}, 20(15): 8413--8422.

\bibitem[{Li et~al.(2021)Li, Wang, Zi, and Zhang}]{transmachine}
Li, J.; Wang, Y.; Zi, Y.; and Zhang, Z. 2021.
\newblock Whitening-Net: A Generalized Network to Diagnose the Faults Among
  Different Machines and Conditions.
\newblock \emph{IEEE Transactions on Neural Networks and Learning Systems},
  1--14.

\bibitem[{Liu et~al.(2020)Liu, Ma, Yan, Wang, Liu, and Ma}]{fedclus}
Liu, Y.; Ma, Z.; Yan, Z.; Wang, Z.; Liu, X.; and Ma, J. 2020.
\newblock Privacy-preserving federated k-means for proactive caching in next
  generation cellular networks.
\newblock \emph{Information Sciences}, 521: 14--31.

\bibitem[{McMahan et~al.(2017)McMahan, Moore, Ramage, Hampson, and
  Arcas}]{fedavg}
McMahan, B.; Moore, E.; Ramage, D.; Hampson, S.; and Arcas, B. A.~y. 2017.
\newblock {Communication-Efficient Learning of Deep Networks from Decentralized
  Data}.
\newblock In Singh, A.; and Zhu, J., eds., \emph{Proceedings of the 20th
  International Conference on Artificial Intelligence and Statistics},
  volume~54 of \emph{Proceedings of Machine Learning Research}, 1273--1282.
  PMLR.

\bibitem[{Sattler, Müller, and Samek(2021)}]{Clustered}
Sattler, F.; Müller, K.-R.; and Samek, W. 2021.
\newblock Clustered Federated Learning: Model-Agnostic Distributed Multitask
  Optimization Under Privacy Constraints.
\newblock \emph{IEEE Transactions on Neural Networks and Learning Systems},
  32(8): 3710--3722.

\bibitem[{Tan et~al.(2021)Tan, Long, Liu, Zhou, Lu, Jiang, and
  Zhang}]{fedproto}
Tan, Y.; Long, G.; Liu, L.; Zhou, T.; Lu, Q.; Jiang, J.; and Zhang, C. 2021.
\newblock FedProto: Federated Prototype Learning over Heterogeneous Devices.
\newblock arXiv:2105.00243.

\bibitem[{Tang, Yuan, and Zhu(2020)}]{CNNreview}
Tang, S.; Yuan, S.; and Zhu, Y. 2020.
\newblock Convolutional Neural Network in Intelligent Fault Diagnosis Toward
  Rotatory Machinery.
\newblock \emph{IEEE Access}, 8: 86510--86519.

\bibitem[{Wang et~al.(2019)Wang, Han, Chu, and Feng}]{VIB}
Wang, T.; Han, Q.; Chu, F.; and Feng, Z. 2019.
\newblock Vibration based condition monitoring and fault diagnosis of wind
  turbine planetary gearbox: A review.
\newblock \emph{Mechanical Systems and Signal Processing}, 126: 662--685.

\bibitem[{Wang et~al.(2021)Wang, Liu, Lin, Chen, Li, Hu, and Chen}]{F1}
Wang, Y.; Liu, R.; Lin, D.; Chen, D.; Li, P.; Hu, Q.; and Chen, C. L.~P. 2021.
\newblock Coarse-to-Fine: Progressive Knowledge Transfer-Based Multitask
  Convolutional Neural Network for Intelligent Large-Scale Fault Diagnosis.
\newblock \emph{IEEE Transactions on Neural Networks and Learning Systems},
  1--14.

\bibitem[{Wang, Wang, and Wang(2018)}]{GAN}
Wang, Z.; Wang, J.; and Wang, Y. 2018.
\newblock An intelligent diagnosis scheme based on generative adversarial
  learning deep neural networks and its application to planetary gearbox fault
  pattern recognition.
\newblock \emph{Neurocomputing}, 310: 213--222.

\bibitem[{Wu, He, and Chen(2020)}]{PerFed}
Wu, Q.; He, K.; and Chen, X. 2020.
\newblock Personalized Federated Learning for Intelligent IoT Applications: A
  Cloud-Edge Based Framework.
\newblock \emph{IEEE Open Journal of the Computer Society}, 1: 35--44.

\bibitem[{Yang et~al.(2020)Yang, Lei, Jia, Li, and Du}]{TL}
Yang, B.; Lei, Y.; Jia, F.; Li, N.; and Du, Z. 2020.
\newblock A Polynomial Kernel Induced Distance Metric to Improve Deep Transfer
  Learning for Fault Diagnosis of Machines.
\newblock \emph{IEEE Transactions on Industrial Electronics}, 67(11):
  9747--9757.

\bibitem[{Zhang et~al.(2021)Zhang, Xu, Gong, Chen, and Gao}]{FLBearing}
Zhang, Z.; Xu, X.; Gong, W.; Chen, Y.; and Gao, H. 2021.
\newblock Efficient federated convolutional neural network with information
  fusion for rolling bearing fault diagnosis.
\newblock \emph{Control Engineering Practice}, 116: 104913.

\bibitem[{Zhu, Chen, and Shen(2020)}]{transfer}
Zhu, J.; Chen, N.; and Shen, C. 2020.
\newblock A New Deep Transfer Learning Method for Bearing Fault Diagnosis Under
  Different Working Conditions.
\newblock \emph{IEEE Sensors Journal}, 20(15): 8394--8402.

\end{thebibliography}

\end{document}